\documentclass{article}
\usepackage{spconf,amsmath,graphicx,amsfonts,mathtools,dsfont,graphicx}
\usepackage{subcaption,amssymb}

\DeclarePairedDelimiter{\norm}{\lVert}{\rVert}

\def\diag{{\text{\bf diag}}}
\newcommand{\boly}{\boldsymbol{y}}
\newcommand{\bolb}{\boldsymbol{\beta}}

\newcommand{\bolh}{\boldsymbol{h}}
\newcommand{\bole}{\boldsymbol{\varepsilon}}
\newcommand{\bolc}{\boldsymbol{c}}
\newcommand{\bolu}{\boldsymbol{u}}
\newcommand{\bolx}{\boldsymbol{x}}
\newcommand{\bolz}{\boldsymbol{z}}
\newcommand{\bolde}{\boldsymbol{\Delta}}

\newcommand{\f}{\tilde f}

\title{Fast Stochastic Hierarchical Bayesian MAP for Tomographic Imaging}
  \name{John McKay$^\dagger$, Raghu G. Raj$^\circledast$, Vishal Monga$^\dagger$\sthanks{Contact:  raghu.raj@nrl.navy.mil.  Funding provided by ONR grant N0001416WX00981.}}
	\address{
	$^\dagger$Pennsylvania State University, University Park, PA\\
	$^\circledast$U.S. Naval Research Laboratory, Washington, DC}

\begin{document}
%
\maketitle
%








\begin{abstract}
Any image recovery algorithm attempts to achieve the highest quality reconstruction in a timely manner. The former can be achieved in several ways, among which are by incorporating Bayesian priors that exploit natural image tendencies to cue in on relevant phenomena. The Hierarchical Bayesian MAP (HB-MAP) is one such approach \cite{raj2016hierarchical} which is known to produce compelling results albeit at a substantial computational cost. We look to provide further analysis and insights into what makes the HB-MAP work. While retaining the proficient nature of HB-MAP's Type-I estimation, we propose a stochastic approximation-based approach to Type-II estimation. The resulting algorithm, fast stochastic HB-MAP (fsHBMAP), takes dramatically fewer operations while retaining high reconstruction quality. We employ our fsHBMAP scheme towards the problem of tomographic imaging and demonstrate that fsHBMAP furnishes promising results when compared to many competing methods.
\end{abstract}
\section{Introduction}
\label{sec:intro}

The act of obtaining images from tomographic settings has been a long studied problem with applications ranging from Radar to magnetic resonance imaging.  However the signaling apparatus is constructed, there is a consistent theme: a plane/volume is captured by rays of directed waves that are recorded, match filtered, and processed to achieve the pixels of the image.  Several aspects to this procedure have emerged as especially difficult, none the least of which is limited sampling.  If we have a tomographic set-up that can collect only a dearth of jittered data points, can we still obtain a quality image?  We look to address this problem.

Of the different manners in which to handle such a situation, the hierarchical Bayesian Maximum a posterior approach (HB-MAP) has yielded compelling state-of-the-art results on optical images with the express intent of usage for, but not limited to, tomography \cite{raj2016hierarchical}.  However this method has one significant drawback: it’s computationally expensive.  Its detailed structure involves a complicated objective function which requires nontrivial computational cost, restricting its applicability to large-scale inverse problems.

In the following, we present a modified HB-MAP algorithm, fast stochastic HB-MAP (fsHBMAP), using a stochastic approximation approach to solving a problematic objective function and provide experimental work to justify its usage in tomographic settings.  In section \ref{sec:HB-MAP}, we present the background structure pertaining to HB-MAP and how its unique construction involves probabilistic priors to allow for robustness in limited sample cases.  Section \ref{sec:modification} outlines our findings considering which aspects to HB-MAP are most influential and how a fast stochastic method can be employed to speed the process up.  Lastly, section \ref{sec:experiments} goes through several tomographic scenarios to illustrate how well fsHBMAP can work.

\section{A Brief Overview of HB-MAP}\label{sec:HB-MAP}

Consider the following linear model for reconstructing images which finds powerful applications in imaging and compressive sensing:  $\boly = X\bolb + \bole$ where $\boly$ a vector of received measurements, $X$ is a dictionary containing some basis, whether it be wavelet, Fourier, or similar, $\bolb$ is the coefficient vector we look to find, and $\bole$ is Gaussian white noise.  We presume $\boly,\bolb,\bole\in\mathbb{R}^n$ and $X\in\mathbb{R}^{n\times n}$.  There is no shortage in ways to solve for $\bolb$ given the assumption of the above linear model and many involve a prior being placed on this coefficient term either explicitly or implicitly via the inclusion of a penalty function \cite{candes2008introduction}.  In this way, it can be further interpreted as solving
\begin{equation}\label{eq:map}
\begin{aligned}
&\arg\max\limits_{\bolb\in\mathbb{F}^n} \log P(\bolb\,|\,\boly)\\
=& \arg\max\limits_{\bolb\in\mathbb{F}^n} \norm{\boly - X\bolb}_2^2 - \log P(\bolb)
\end{aligned}
\end{equation}

What is the ``best" choice for the prior on $\bolb$?  In the compressive sensing literature a popular choice for this distribution is the Laplacian i.e. $log P(\bolb) \propto -\norm{\bolb}_1$, which can be interpreted as a relaxation of sparsity-enforcing $\ell_0$ pseudo norm penalty \cite{babacan2010bayesian}.  While popular, this choice for $P(\bolb)$ lacks proper sophistication in order to capture certain natural phenomena \cite{raj2016hierarchical,mousavi2015iterative}.  This is where the HB-MAP comes into play.

Before we start, let us first revisit our linear model construction.  For HB-MAP, we will think of $X$ as the product of two quantities: a measurement matrix $\Psi$ and a dictionary $\Phi$.  With $X=\Psi\Phi$, we still have $\bolb$ as a coefficient vector except with the true image being $\Phi\bolb$.  This grants us the flexibility to further incorporate convenient linear transformation spaces (typically, but not restricted to, wavelet) concerning our image $I$.  $\Psi$ could hold the information regarding the arrangement of sensors and readings, primarily in the mold of a Radon transform matrix in the tomographic setting \cite{ramm1996radon}, and $\Phi$ a wavelet dictionary such that $\Phi\bolb$ is the optical image we are looking to recover.  What then is a suitable prior for $\bolb$ since it is now the wavelet coefficients of a natural image?  The work of \cite{raj2016hierarchical,simoncelli1997statistical,wainwright2000random} suggest that it may be a compound Gaussian.

One can dive deeply into the details of compound Gaussians by addressing the work of \cite{raj2016hierarchical,simoncelli1997statistical,wainwright2000random,mallat1999wavelet}, among others, but the most basic understanding necessary for HB-MAP boils down to this:  $\bolb$ can be though of as
\begin{equation}\label{eq:beta}
\bolb = \bolu\odot\bolz \text{ where }\bolu\sim\mathcal{N}(0,\Sigma_u) \text{ and }\bolz = \bolh(\bolx)
\end{equation}
with $\bolx$ following some mutli-scale Gaussian tree structure and $\bolh(\cdot)$ being an entry-wise, usually nonlinear function.  This construction leads to a two stage solution starting with a calculation for $\bolz$ (Type-II estimation) followed by a scheme for $\bolu$ (Type-I estimation).

The Type-II estimation involves substantial computations and is the main focus of our proposed improvement.  The issues arise from $\bolx$; we can state $\boly$ as
\begin{equation}\label{eq:linearModelodot}
\boly = X(\bolh(\bolx)\odot \bolu)+\bole
\end{equation}
then when we look to optimize $\bolz$, that really is the same as optimizing $\bolx$ as the input to the nonlinear $\bolh$.  \cite{raj2016hierarchical} demonstrates in detail as to how Type-II estimation, due to the multi-scale Gaussian hierarchical Bayesian structure imposed on $x$, comes down to solving
\begin{equation}\label{eq:f}
\begin{aligned}
\arg\max\limits_{\bolx\in\mathbb{R}^n} \boly^T B(\bolx)^{-1} \boly + \log\det B(\bolx) + \bolx^T\Sigma_x^{-1}\bolx
\end{aligned}
\end{equation} 
\begin{equation}\label{eq:B}
B(\bolx)=XH(\bolx)\Sigma_uH(\bolx)X^T+\Sigma_\epsilon
\end{equation}
given $H(\bolx)=\diag(\bolh(\bolx))$.  For this objective function, which we will define as $f(\bolx)$, \cite{raj2016hierarchical} suggests an analytically straightforward steepest descent method.  Unfortunately, $f(x)$ has what would be best described as a particularly nasty gradient in terms of computational efficiency.  Even with a ``nice" choice for $\bolh$ (in terms analytical properties such as differentiability), the gradient, the gradient requires costly Kronecker products which are unavoidable without severe modification.  We delve into this matter more in section \ref{sec:modification} where we demonstrate how to circumvent the calculation of an unseemly $\nabla_x f$ and even more complex $\nabla_x^2 f$ (both of which are detailed in \cite{raj2016hierarchical}.

Assuming that we obtain an optimal $\bolx^\ast$, Type-I estimation looks to build off of this to calculate $\bolu$.  This in turn is much simpler as
\begin{equation}
\begin{aligned}
\bolu^\ast &= \arg\max\limits_{\bolu\in\mathbb{F}^n} \log P(\boly\,|\,\bolu) +\log P(\bolu)\\
&= \arg\min\limits_{\bolu\in\mathbb{F}^n} (\boly-XH(\bolx^\ast)\bolu)^T\Sigma_\varepsilon^{-1}(\boly-XH(\bolx^\ast)\bolu)\\
&\,\,\,\,\,\,\,\,\,\,\,\,\,\,\,\,\,\,\,\,\,\,\,\,\,\,\,\,\,\,\,\,\,\,\,\,\,\,\,\,
+\bolu^T\Sigma_u^{-1}\bolu
\end{aligned}
\end{equation}
which, for $\Lambda=\diag(\bolz^\ast)$, leads to
\begin{equation}\label{eq:usolution}
\Lambda (X^T\Sigma_\varepsilon^{-1}X+\Lambda^{-1}\Sigma_u\Lambda^{-1})\Lambda\bolu = \Lambda X^T\Sigma_\varepsilon^{-1}\boly
\end{equation}
Since this computation for $\bolu$ can be nontrivial in the manner given by \eqref{eq:usolution}, \cite{raj2016hierarchical} employed a strategy to reduce the load; defining $\Lambda_\tau$ as
\begin{equation}
\Lambda_\tau = \diag(\bolc) \text{ where } c_i=
\begin{cases}
1, & z_i^\ast>\tau\\
0, & \text{ else}
\end{cases}
\end{equation}
and then finding $\bolu$ with
\begin{equation}
\Lambda_\tau (X^T\Sigma_\varepsilon^{-1}X+\Lambda^{-1}\Sigma_u\Lambda^{-1})\Lambda_\tau\bolu = \Lambda X^T\Sigma_\varepsilon^{-1}\boly
\end{equation}

With both $\bolz=h(\bolx^\ast)$ and $\bolu$, one can quickly find $\bolb$ using \eqref{eq:beta}, completing the two step process.

\section{fast stochastic HB-MAP}\label{sec:modification}

Before we dive into our handling of the Type-II estimation aspect of HB-MAP, we first wish to provide context for the reader concerning what is happening within the HB-MAP algorithm.  That is, what improvements can be had concerning Type-II estimation?  To what degree can Type-I alone handle the problem of image reconstruction?  Our research has surprisingly shown that the answer to the latter question is:  quite a bit.

Let us revisit \eqref{eq:usolution}; here if we do not do any simplifying substitutions regarding $\Lambda$ then we can reduce the problem to
\begin{equation}\label{eq:usol1}
(\tfrac{1}{\sigma_\varepsilon^2}X^TX+\tfrac{1}{\sigma_u^2}\Lambda^{-2})\Lambda\bolu=\tfrac{1}{\sigma_\varepsilon^2}X^T\boly
\end{equation}
with $\Sigma_\varepsilon=\sigma^2_\varepsilon I$ and $\Sigma_u=\sigma_u^2I$.  Notice that 
\begin{equation}
\Lambda\bolu=\bolz\odot\bolu=\bolb
\end{equation}
\begin{equation}\label{eq:bolbzu}
\implies\bolb=\tfrac{1}{\sigma_\varepsilon^2}(\tfrac{1}{\sigma_\varepsilon^2}X^TX+\tfrac{1}{\sigma_u^2}\Lambda^{-2})^{-1}X^T\boly
\end{equation}
The optimality of $\bolb$ hinges on the assumed optimality of $\bolx^\ast$ (i.e. where $h(\bolx^\ast)$ constitutes the diagonal of $\Lambda$) - but - what if we used another suboptimal $\bolx$ as the input to $\bolh$?

\begin{figure}[t]\centering
\includegraphics[width=1\columnwidth]{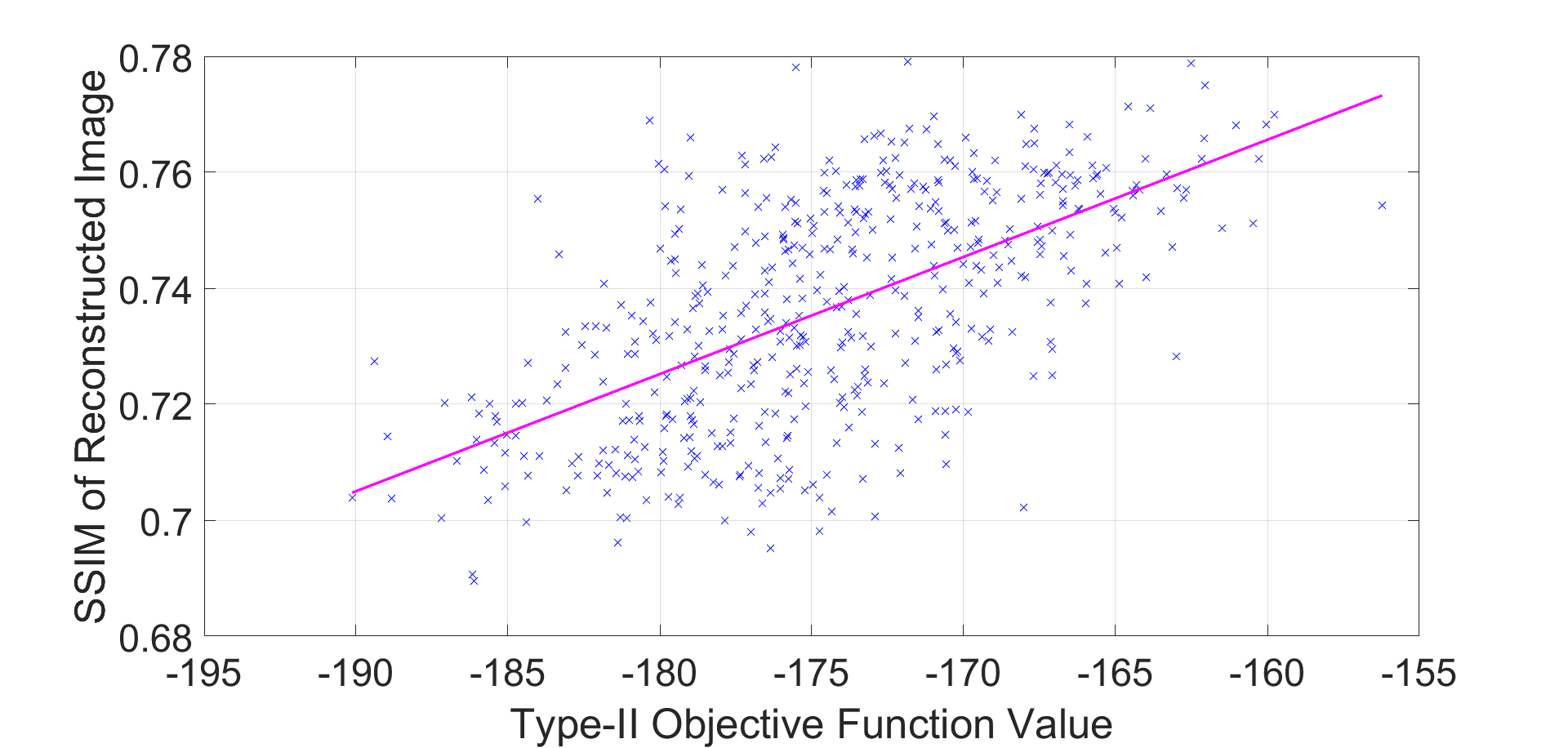}
\caption{Comparison of 400 randomly chosen $\bolx$ obj. function\eqref{eq:f} value \& vs. subsequent reconstructed image SSIM.  Each test involved a radon transform on a $16\times 16$ snippet of Barbara image.  The trend line is provided for context.}\label{fig:testValues}
\end{figure}

As it turns out, if the value of $\sigma_\varepsilon^2$ is known or, more likely, estimated well, then $\tilde \bolb$, the coefficient vector corresponding to the solution of \eqref{eq:bolbzu} using a suboptimal $\bolx$, can actually perform non-trivially.  As evidence to such, consider the scenario illustrated by Figure \ref{fig:testValues}; here, we conducted 1000 image recounstructions of a $16\times 16$ snippet of the Barbara image in the form of an inverse radon transform with added noise (the sampling pattern of the radon transform can be seen in Figure \ref{sub:a}, the noise was $\sigma^2_\varepsilon=.1$, and a wavelet dictionary was used).  In each case, a vector composed of uniformly random values within $(0,1)$ made up $\bolx$ and no refinement was made.  Notice that this relatively simple procedure yielded SSIM \cite{wang2004image} metrics with a mean of .73 and entirely within the range of $(.68,.78)$.

Given this aspect of HB-MAP, the question then becomes of what influence the Type-II estimation has over the algorithm's output.  We reference again Figure \ref{fig:testValues}; the trend line points to the fact that maximized values for $f$ lead to better images.  Thus, given any starting point, it appears useful to try to transition towards higher Type-II objective function values.  We propose how to do so.

As mentioned in section \ref{sec:HB-MAP}, the Type-II estimation proposed by \cite{raj2016hierarchical} involves a prohibitively costly gradient calculation (to say nothing of its Hessian).  How to handle such a problem?  We will outline here an approximation $\f$ that is much more efficient and, as we show in section \ref{sec:experiments}, still impressively capable.  To start, let us break $f$ into:
\begin{equation}\label{eq:parts}
f(\bolx)=\underbrace{\boly^T B(\bolx)^{-1} \boly}_{f_1(\bolx)} + \underbrace{\log\det B(\bolx)}_{f_2(\bolx)} + \underbrace{\bolx^T\Sigma_x^{-1}\bolx}_{f_3(\bolx)}
\end{equation}
Just by inspection, a reader with some familiarity regarding numerical optimization may see that $f_3$ represents a more normal-looking objective function.  The key technical issues revolve around ameliorating the numerical tractability of $f_1$ and $f_2$ given their role in $B(x)$ as defined in \eqref{eq:B}.

Let's briefly start right there with $B(\bolx)$.  Following \cite{raj2016hierarchical}, we use the non-linear choice of $\bolh$ 
\begin{equation}\label{eq:h}
\bolh(\bolx) = \left[\sqrt{\exp(\tfrac{x_1}{a})}\cdots\sqrt{\exp(\tfrac{x_n}{a})}\right]^T
\end{equation}
with $a\in\mathbb{R}$ (note that our framework can use a general choice for $bolh$).  Substituting this and our other assumptions for the matrices $\Sigma_u$ and $\Sigma_\varepsilon$ into \eqref{eq:B} yields
\begin{equation}
B(\bolx) = \sigma_u^2X^TH^2(\bolx)X+\sigma_\varepsilon^2I
\end{equation}

Introducing a slightly simplified $B(\bolx)$ and formal choice for $H$ is necessary for our first challenge:  $f_2$.  Before we start, we want to make two more assumptions:  $\Psi$ is square and full rank.  Now, this may not be the case in practice - but - the proceeding theory provides a strong basis from which we shown in later sections is justified by experimental work.  With all that said, we first point out that, by the Minkowski inequality
\begin{equation}\begin{aligned}
\det(B(\bolx))=&\det(\sigma_u^2X^TH^2(\bolx)X+\sigma_\varepsilon^2I)\\
&\geq \det(\sigma_u^2X^TH^2(\bolx)X)+\det(\sigma_\varepsilon^2I)
\end{aligned}\end{equation}
Now, consider the determinant of the matrix $X$;  The dictionary $D$ is typically chosen so that it is unitary, meaning that $\det(D)=1$.  Thus, we see that, $\det(X)=\det(RD)=\det(R)$.  Therefore, since $\log$ is monotonically increasing,
\begin{equation}
\det(\sigma_u^2X^TH^2(\bolx)X)=\sigma_u^{2n}\det(R)^2\det(H^2(\bolx))
\end{equation}
Which, returning to the $\log\det$ function means
\begin{equation}\begin{aligned}
\log\det(B(\bolx))\geq & \log(\sigma_u^{2n}\det(R)^2\det(H^2(\bolx))+\sigma_\varepsilon^{2n})\\
\geq & \log\det(H^2(\bolx))+K\\
= & \sum\limits_{i=1}^n\frac{x_i}{a}+K
\end{aligned}\end{equation}
for $K=\log(\sigma_u^{2n}\det(R)^2)+\log(\sigma_\varepsilon^{2n})$.  We define an approximation for $f_2$ as $\tilde f_2(\bolx)=\sum_{i=1}^{n}\tfrac{x_i}{a}$.

\begin{figure}\centering
\includegraphics[width=\columnwidth]{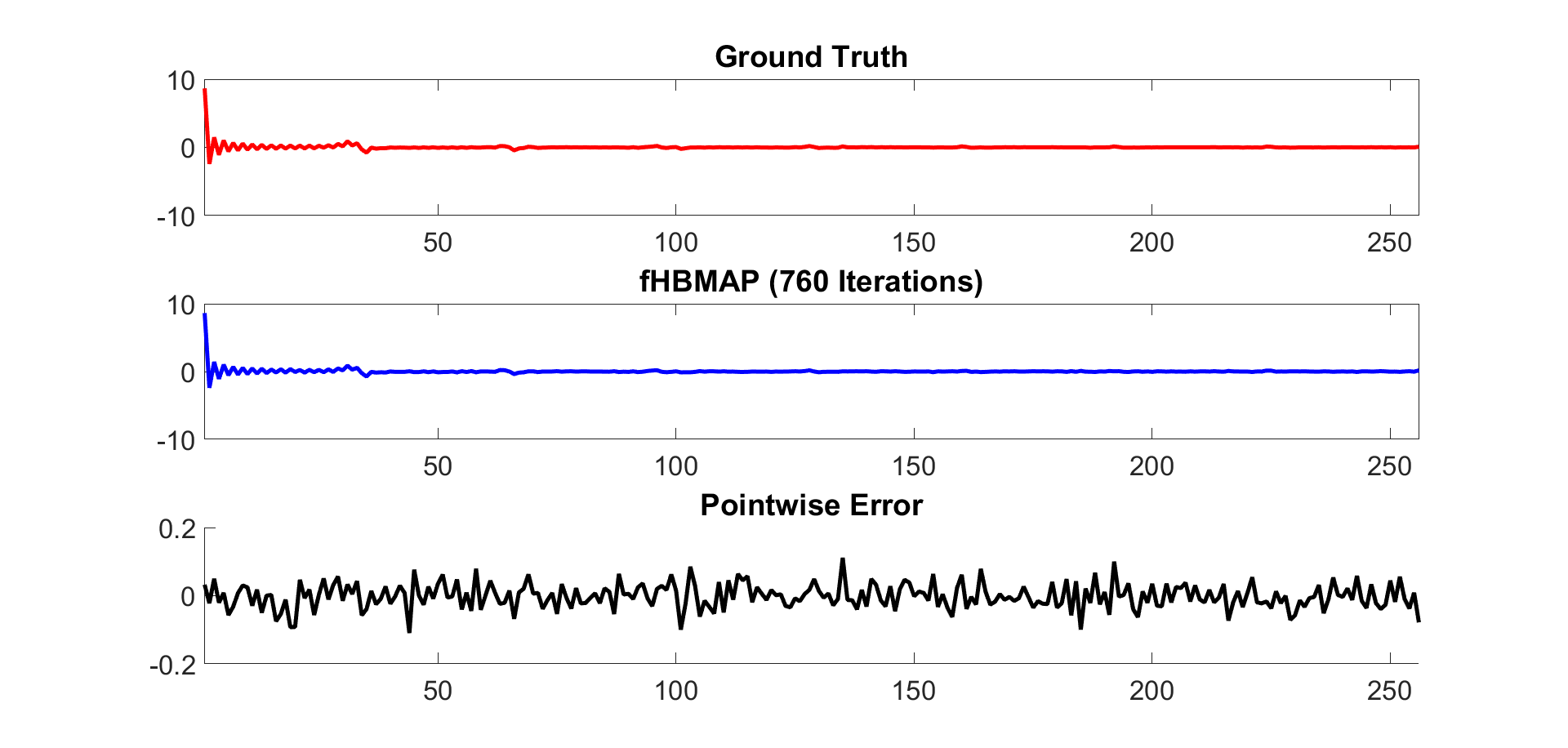}
\caption{Comparison of wavelet coefficients of action (top) \& fsHBMAP (middle) with point-wise error given (bottom).}\label{fig:coef}
\end{figure}

Moving onto $f_1$ reveals no readily actionable approximation.  The inverse of $B(\bolx)$, which is composed of the sum of matrices, is a highly nontrivial problem.  In \cite{raj2016hierarchical} we attacked this problem by a head-on approch which involved very expensive gradient and/or Hessian calculations which, while furnishing exact results, thereby limit its direct applicability to large-scale inverse problems. The path forward, as we see it, is to avoid gradient calculations altogether by invoking a stochastic approximation (SA) approach.

\begin{figure}[t]\centering
\includegraphics[width=1\columnwidth]{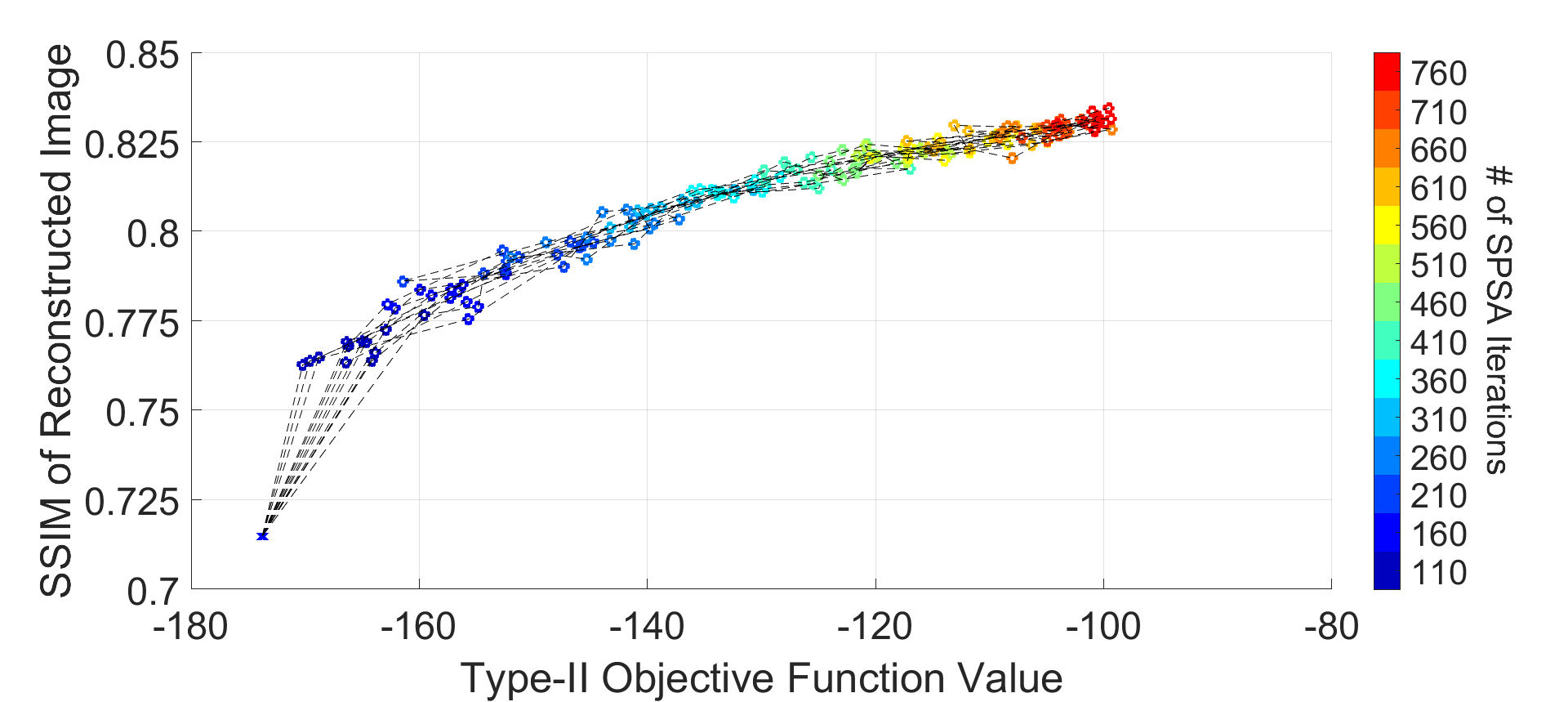}\\
\includegraphics[width=1\columnwidth]{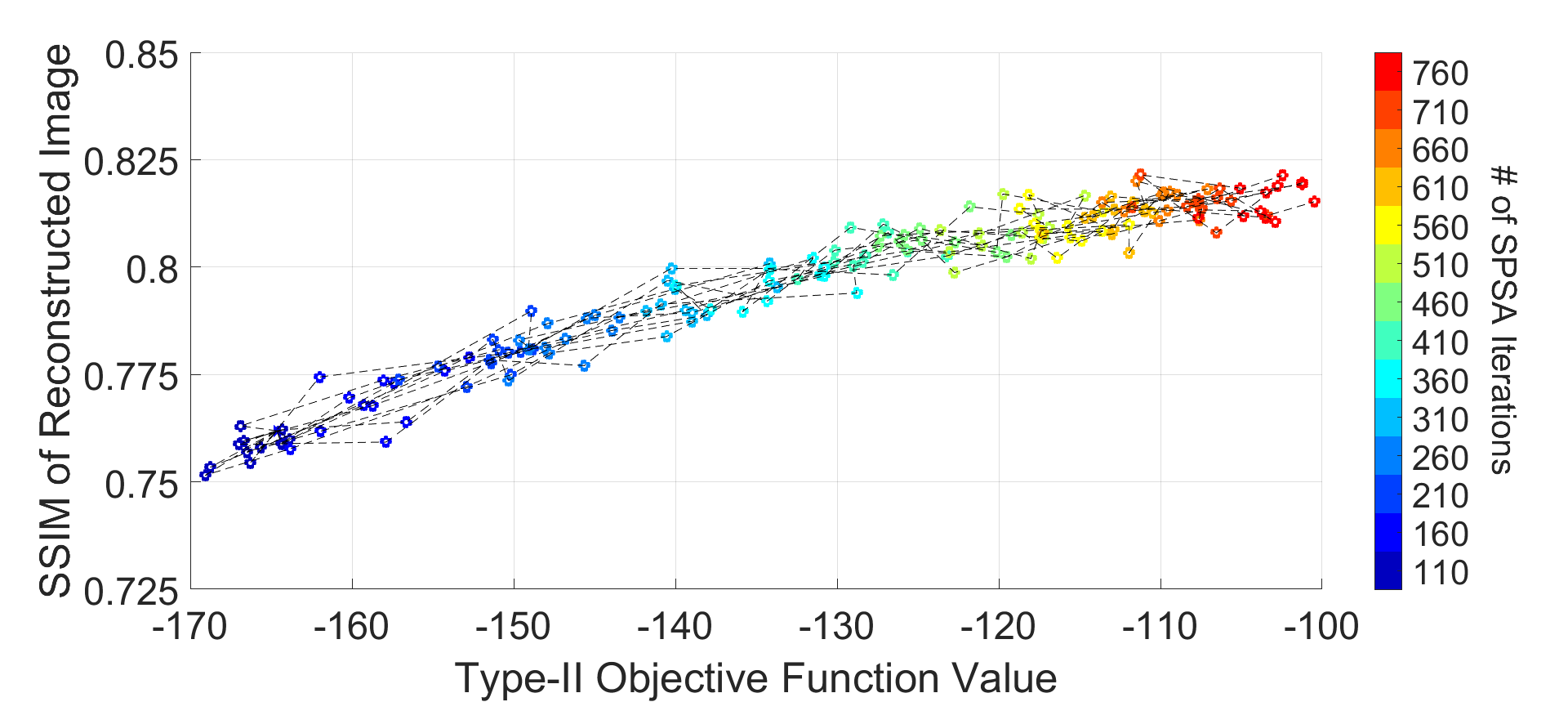}\\
\begin{tabular}{r|c|c|c|c|c|c|c|c|c|}
\textbf{Iteration}&260&360&460&560&660&760\\\hline
\textbf{Time (s)}&4.72&6.52&8.42&10.55&13.26&13.85\\\hline
\end{tabular}
\caption{Using setup of Fig. \ref{fig:testValues} we took random initial values \& evaluated SSIM values \cite{wang2004image} for varied SPSA iterations using wavelet (top) and DCT (bottom) dictionaries.  Mean fsHBMAP completion times (seconds) are given (wavelet).}\label{fig:upswing}
\end{figure}

If we were to start with the idea that we just wanted a near-enough calculation of the gradient for use in a steepest descent technique, then it would not be unreasonable to pursue some finite difference scheme.  SA builds off this idea; the crux of methods like finite difference SA (FDSA) and simultaneous perturbation SA (SPSA) start with finite difference ideas and incorporate probabilistic theories to reduce computations \cite{spall1992multivariate,spall1998overview,glynn1986stochastic}.  For SPSA, the idea is to modify the update step of a gradient descent algorithm for step size $\alpha_k$
\begin{equation}
\bolx_{k+1}=\bolx_k-\alpha_k\tilde g(\bolx)
\end{equation}
so that there is a gradient approximation, $\tilde g$
\begin{equation}
\tilde g(\bolx) =\tfrac{1}{c_k}(f(\bolx+c_k\bolde_k)-f(\bolx-c_k\bolde_k)) \oslash \bolde_k
\end{equation}
where $c_k>0$ is a small value that decreases as $k\to\infty$, $\bolde_k$ follows a symmetric $\pm 1$ Bernoulli distribution, and $\oslash$ is the element-wise division operator.  Given the smoothness of $f$ and sufficient choices for $\alpha_k$ and $c_k$, we found that SPSA served as competent method for Type-II estimation with substantially fewer calculations.  Note that SPSA involves only \emph{two} evaluations of the loss function at each iteration.

To gain an idea of how helpful SPSA can be and what computation effort can be expected, refer to Figure \ref{fig:testValues}.  Here we used the same $16\times 16$ snippet and radon transform as previous and tried over ten trials increasing numbers of SPSA iterations with wavelet and DCT dictionaries.  Random initializations started SPSA (both in Figure 3 and in the experiments in Section 4).  In every case, SPSA was not only able to decrease the value of $f$ but also do so in a way that improves the ultimate SSIM value of the reconstructed image.  SPSA was able to replicate the sparse signal from the randomized starting point as Figure \ref{fig:coef} shows.


\section{Experiments}\label{sec:experiments}
\def\sep{.6}
\def\siz{1}
\begin{figure*}[t]
\centering
\begin{subfigure}{\sep\columnwidth}\centering
\includegraphics[width=1\columnwidth]{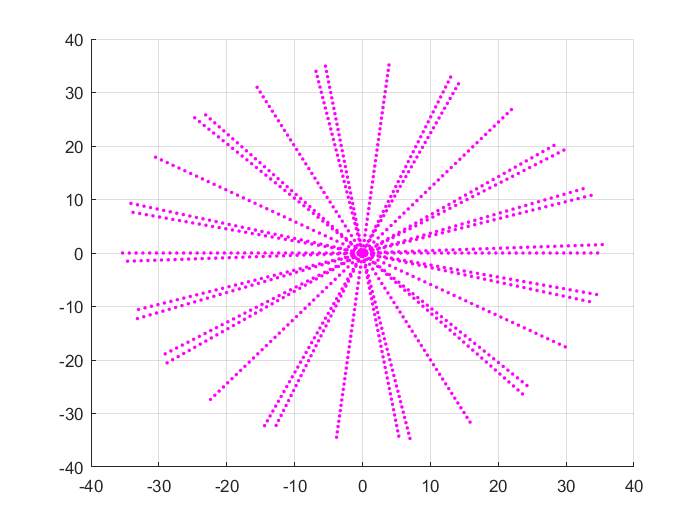}
\caption{Projection}%
\label{sub:a}%
\end{subfigure}
\begin{subfigure}{\sep\columnwidth}\centering
\includegraphics[width=\siz\columnwidth]{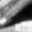}
\caption{Orig. (SSIM)}%
\label{sub:b}%
\end{subfigure}
\begin{subfigure}{\sep\columnwidth}\centering
\includegraphics[width=\siz\columnwidth]{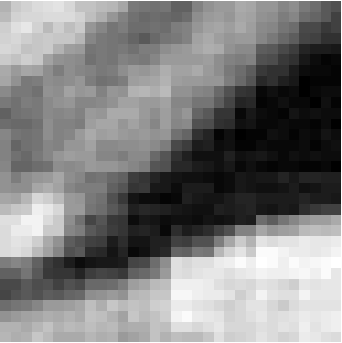}
\caption{gHBMAP \cite{raj2016hierarchical} (.9)}%
\label{sub:cc}%
\end{subfigure}
\begin{subfigure}{\sep\columnwidth}\centering
\includegraphics[width=\siz\columnwidth]{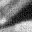}
\caption{fsHBMAP(.78)}%
\label{sub:d}%
\end{subfigure}
\begin{subfigure}{\sep\columnwidth}\centering
\includegraphics[width=\siz\columnwidth]{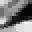}
\caption{CoSaMP \cite{needell2009cosamp} (.74)}%
\label{sub:e}%
\end{subfigure}
\begin{subfigure}{\sep\columnwidth}\centering
\includegraphics[width=\siz\columnwidth]{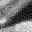}
\caption{ICR \cite{mousavi2015iterative} (.73)}%
\label{sub:g}%
\end{subfigure}
\begin{tabular}{r|c|c|c|c|}
&(c) gHBMAP&(d) fsHBMAP&(e) CoSaMP&(f) ICR\\\hline
\textbf{Time (seconds)}&480&54&3&61\\\hline
\end{tabular}
\caption{Reconstructions of $32\times 32$ Barbara photo snippet after projected, sampled in the pattern of \ref{sub:a}.  Noise was $\sigma_\varepsilon^2=.1$.  Newton HB-MAP had an SSIM of .94 at a cost of $1.1\times 10^{4}$ seconds (not shown for space).}\label{fig:radonmatrix}
\end{figure*}

To provide context for fsHBMAP, we experimented on a $32\times 32$ piece of the Barbara image projected with the sampling pattern show in Figure \ref{sub:a} which contains 18 rays containing 71 samples, each.  Overall, the gradient HB-MAP reconstruction (gHBMAP) \cite{raj2016hierarchical} was best in terms of SSIM with a value of .9. fsHBMAP is the next best method with a SSIM of .78 followed by CoSaMP \cite{needell2009cosamp} (.74), iterative convex refinement \cite{mousavi2015iterative} (.73), and OMP (.72). While fsHBMAP did not achieve the quality of the HB-MAPs, it still outperforms all other competing methods but with a substantially less computation time. In particular, fsHBMAP was able to achieve its reconstruction in 4\% of the time it took gHBMAP! Furthermore fsHBMAP is much more amenable to being applied to large scale imaging problems because--unlike gHBMAP--no explicit gradient or Kronecker calculations are involved.  In the case of Newton HB-MAP (nHBMAP) proposed by \cite{raj2016hierarchical}, while it achieves the best SSIM value, it computational expense makes it unsuitable for most practical applications.

We have thus offered a natural-image inspired method for image reconstruction.  Future work includes the application of second order SPSA algorithms to fsHBMAP and/or more computationally sophisticated ways to make larger image scenes more manageable.


%
%
%
%
%
%





\bibliographystyle{IEEEbib}
\bibliography{refGlobalSIP2017}\label{sec:refs}

\end{document}